\documentclass{article}

\usepackage[margin=3cm, bottom=3cm, footskip=1.5cm]{geometry}

\usepackage{amsmath,amssymb,amsthm}
\usepackage{algorithm2e}
\usepackage[subrefformat=parens]{subcaption}
\usepackage[bbgreekl]{mathbbol}
\usepackage{graphicx}
\usepackage{bm}
\usepackage{url}
\usepackage{natbib}
\usepackage{booktabs}

\usepackage{cuted}
\usepackage{adforn}

\title{Real-time tree search with pessimistic scenarios}
\author{Takayuki Osogami \and Toshihiro Takahashi}
\date{}

\begin{document}

\maketitle

\begin{abstract}
Autonomous agents need to make decisions in a sequential manner, under
partially observable environment, and in consideration of how other
agents behave.  In critical situations, such decisions need to be made
in real time for example to avoid collisions and recover to safe
conditions.  We propose a technique of tree search where a
deterministic and pessimistic scenario is used after a specified
depth.  Because there is no branching with the deterministic scenario,
the proposed technique allows us to take into account the events that
can occur far ahead in the future.  The effectiveness of the proposed
technique is demonstrated in Pommerman, a multi-agent environment used
in a NeurIPS 2018 competition, where the agents that implement the
proposed technique have won the first and third places.
\end{abstract}

\section{Introduction}

Autonomous agents, such as self-driving cars and drones, need to make
decisions in real time (under tight time constraints), which is
particularly important but difficult in critical situations for
example to avoid collisions.  Such decisions often need to be made in
a sequential manner to achieve the eventual goal ({\it e.g.}, avoiding
collisions and recovering to safe conditions), under partially
observable environment, and by taking into account how other agents
behave.  Towards this far-reaching goal of realizing such autonomous
agents, we propose practical techniques of sequential decision making
in real time and demonstrate their effectiveness in Pommerman, a
multi-agent environment that has been used in one of the competitions
held at the Thirty-second Conference on Neural Information Processing
Systems (NeurIPS 2018) on Dec.~8, 2018 \citep{pommerman}.  The
techniques that we propose in this paper have been used in the
Pommerman agents (HakozakiJunctions and dypm-final) who have won the
first and third places in the competition.

\begin{figure}[t]
  \begin{minipage}{0.49\linewidth}
    \centering
    \includegraphics[width=\linewidth]{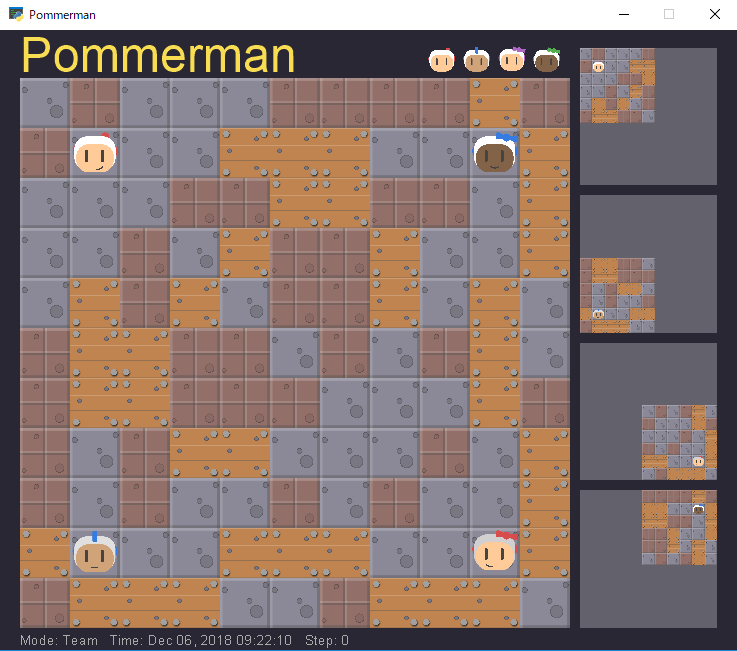}\\
    (a) Initial board
  \end{minipage}
  \begin{minipage}{0.49\linewidth}
    \centering
    \includegraphics[width=\linewidth]{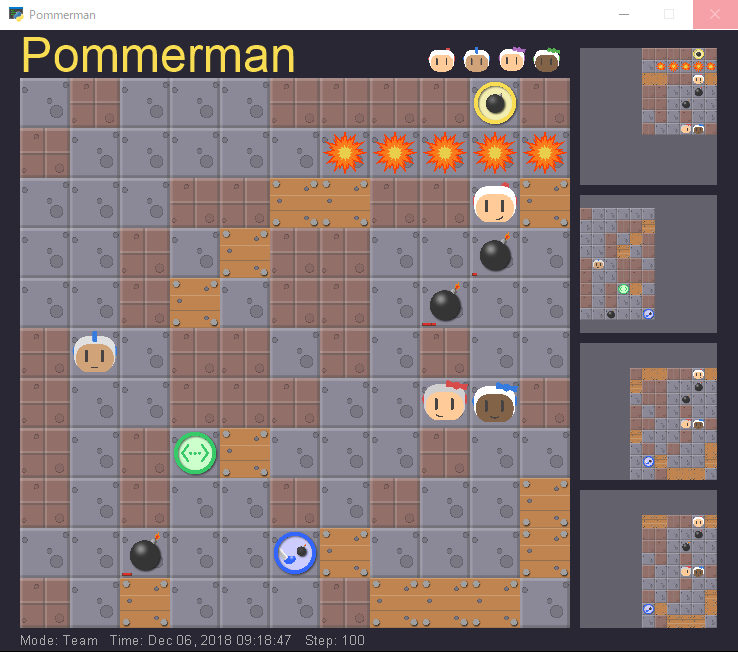}\\
    (b) Board after 100 steps
  \end{minipage}
  \caption{An initial board (a) and a board after 100 steps in
    Pommerman.  The four small windows on the right most column
    respectively denote the areas that the four agents can observe.}
  \label{fig:pommerman}
\end{figure}

In Pommerman,
a team of two agents competes against another team of two agents on a
board of $11\times 11$ grids (see Figure~\ref{fig:pommerman}~(a) for
an initial configuration of the board).  Each agent can observe only a
limited area of the board, and the agents cannot communicate with each
other.  The goal of a team is to knock down all of the opponents.
  Towards this goal, the agents
place bombs to destroy wooden walls and collect power-up items that
might appear from those wooden walls, while avoiding flames and
attacking opponents.  See Figure~\ref{fig:pommerman}~(b) for an example
of the board in the middle of the game.  See \citet{pommerman} and the
GitHub repository\footnote{
  https://github.com/MultiAgentLearning/playground } for details of
Pommerman.

Although Pommerman has been developed recently (initial GitHub commit
was Dec.~25, 2017), it has been gaining much attention as a benchmark
of multi-agent study in the field of planning, game theory, and
reinforcement learning \citep{MARLsurvey2018a,MARLsurvey2018b}.  Prior
to the one at NeurIPS 2018, the first competition was held on Jun.~3,
2018.  The winning agent sets an intermediate goal with heuristics and
performs depth-limited tree search to achieve that intermediate goal
\citep{Zhou18}.  \citet{backplay} propose a technique of imitation
learning with curriculum and show that, by using the behavioral data
of the winning agent, it can train an agent as strong as the winning
agent.

While the community has made progress in developing strong agents for
Pommerman, the level of those agents is not yet comparable to what has
been achieved for backgammon \citep{backgammon}, Chess
\citep{DeepBlue}, Atari video games \citep{DQN}, Poker
\citep{Bowling15,BroSan19}, and Go \citep{AlphaZero}.  Pommerman has
its own difficulty that prohibits effective applications of existing
approaches that have seen success in other games.  For example,
although deep reinforcement learning \citep{DQN} and planning methods
\citep{LRG15} have seen success on Atari video games, they rely on the
simulators of the video games at the phase of either learning or
planning.  These methods cannot be directly applied to (or less
effective in) Pommerman, where opponents are not known in advance and
cannot be simulated.

What makes Pommerman difficult is the constraint on real-time decision
making ({\it i.e.}, an agent needs to choose an action in 100
milliseconds).  This tight time constraint significantly limits the
applicability of Monte Carlo Tree Search, which would otherwise be a
reasonable approach to Pommerman \citep{matiisen2018pommerman}.  In
Pommerman, the branching factor at each step can be as large as
$6^4=1,296$, because four agents take actions simultaneously in each
step, and there are six possible actions for each agent.  The agents
should plan ahead and choose actions by taking into account the
explosion of bombs, whose lifetime is 10 steps.  Tree search with
insufficient depth (less than 10) would ignore the explosion of bombs,
which in turn would make the agents easily caught up in flames.  Tree
search with sufficient depth (at least 10) is practically infeasible
with the large branching factor.  Other difficulties of Pommerman
include the following.  Reward is only given at the end of an episode,
which can be as long as 800 steps.  The agents can observe only a
limited part of the board, and some of the key information cannot be
directly observed.  The agent needs to coordinate with its teammate
without explicit communication.

Here, we propose a practical approach to real-time tree search that
allows us to take into account critical events that can occur far
ahead in the future.  In our approach, tree search after a specified
depth is performed under the assumption of a deterministic and
pessimistic scenario ({\it i.e.}, sequence of states).  Because the
scenario is deterministic, there is no branching after the specified
depth, which allows us to perform the tree search with sufficient
depth to take into account the critical events 
in the distant future.  This deterministic scenario is designed to be
pessimistic by allowing multiple unfavorable events can happen ({\it
  e.g.}, by letting opponents take multiple actions) simultaneously in
a nondeterministic manner.  Hence, our pessimistic scenarios are
unrealistic in general.  Our key idea is that an unrealistic scenario
can capture critical events in the future better than a small number
of realistic scenarios that can be sampled and explored under the
tight time constraint.  We adjust the level of pessimism via self-play
to achieve the best overall performance.

Our approach is proposed particularly for Pommerman but can be
generally applicable to other domains that require real-time
sequential decision making under tight time constraints.  We
demonstrate the flexibility of the proposed approach by instantiating
it as two variants of Pommerman agents, who need to deal with the
complex environment that involves multiple agents and partial
observability.  The effectiveness of the proposed approach is shown
with Pommerman.  The new approach of real-time tree search with
deterministic and pessimistic scenarios and its application to
Pommerman constitute the contributions of this paper.

\section{Related Work}

There has been a significant amount of work on the techniques of tree
search for real-time (strategy) games.  As we will discuss it in the
following, however, the focus of the prior work is on the techniques
for reducing the search space or guiding the search towards the most
relevant subspace.  The novelty in our approach is in synthesizing the
deterministic and pessimistic scenarios.

The prior work has investigated various techniques to make Monte Carlo
Tree Search (MCTS) applicable to real-time games such as Ms.\ Pac-Man
\citep{MCTS-pacman,IkeIto11}, StarCraft \citep{MCTS-StarCraft}, Wargus
\citep{UCT-RTS}, Physical Traveling Salesman Problem
\citep{MCTS-TSP}, Quantified Constraint Satisfaction Problem
\citep{MCTS-QCSP}, and $\mu$RTS \citep{Puppet2,Mar19}.  An example of a
recent work in this line is \citet{Mar19}, who study a technique of
action abstraction and apply it to MCTS among others to reduce the
search space.  \citet{Puppet2} study a technique of using
non-deterministic rules to reduce the branching in MCTS.  In all of
such prior work, MCTS is performed only with realistic or legal moves.

\citet{MTCS-Atari} study the approach of using MCTS to generate
training data for learning a deep neural network that approximates a
policy or a value function, and the effectiveness of the proposed
approach is demonstrated in Atari video games.  This approach is motivated
by the observation that tree search (planning-based approaches) can
perform far better than model-free approaches if it were not for the
tight time constraint.



Real-time tree search has also been studied for deterministic
settings.  Here, the search tree is expanded on the basis of heuristic
values as long as time permits.  Similar to real-time MCTS, key
questions are where to expand and what actions to take given the
search tree investigated.  For example, \citet{Mit19} propose a
risk-sensitive approach to these questions.

Our approach of synthesizing a pessimistic scenario is also related to
the null-move heuristic, which has been studied particularly for Chess
\citep{nullmove}, in that it considers a scenario with illegal moves.
The null-move heuristic assumes that a player skips a move, which is
illegal in chess, to estimate a lower bound of the value of the best
move.  The lower bound is then used to prune the search space.  In
contrast, our approach can assume that a player takes multiple moves in
the pessimistic scenario.

A pessimistic scenario is also similar to delete relaxation or relaxed
plan heuristics \citep{HofNeb01} in classical planning.  Delete
relaxation is similar to a pessimistic scenario in the sense that it
does not ``delete'' an opponent from a position.  A difference is that
a pessimistic scenario allows an opponent to take multiple actions
simultaneously.

Our approach is also similar to ``variable resolution''
\citep{VariableResolution} in that exact search is limited to a
certain depth.  After that depth, the two approaches differ.  In
particular, a pessimistic scenario cannot be obtained by ``removing
some information from the planning task,'' as is done in
\citet{VariableResolution}.


\section{Real-time tree search for Pommerman}
\label{sec:tree}

In Pommerman, the dynamics of the environment is known, and much of
the uncertainties resulting from partial observability can be resolved
with careful analysis of historical observations.  MCTS would thus be
a competitive approach if it were not for the tight time constraint \citep{matiisen2018pommerman}.
For example, consider a situation where an agent can survive only by
following a particular route.  Tree search is particularly suitable
for finding such a route, while model-free approaches of learning
policies or value functions, if not impossible, would require large scale
functional approximators ({\it e.g.}, deep neural networks)
and a large amount of data for training to be able to follow that
route.  The applicability of MCTS or tree search in general is however
significantly limited in Pommerman due to the tight time constraint and
the large branching factor.

One approach of tree search is to push the depth as far as possible,
and this is the approach taken by the gorogm\_eisenach ({\tt
  eisenach}) agents, who won the second place in the NeurIPS 2018
Pommerman competition.  The {\tt eisenach} agent was implemented in
C++ with various engineering tricks to achieve the average depth of 2
in the tree search \citep{PommermanNeurIPS2018}.

\subsection{Tree search with pessimistic scenarios}

\begin{figure}
  \centering
  \includegraphics[width=0.66\linewidth]{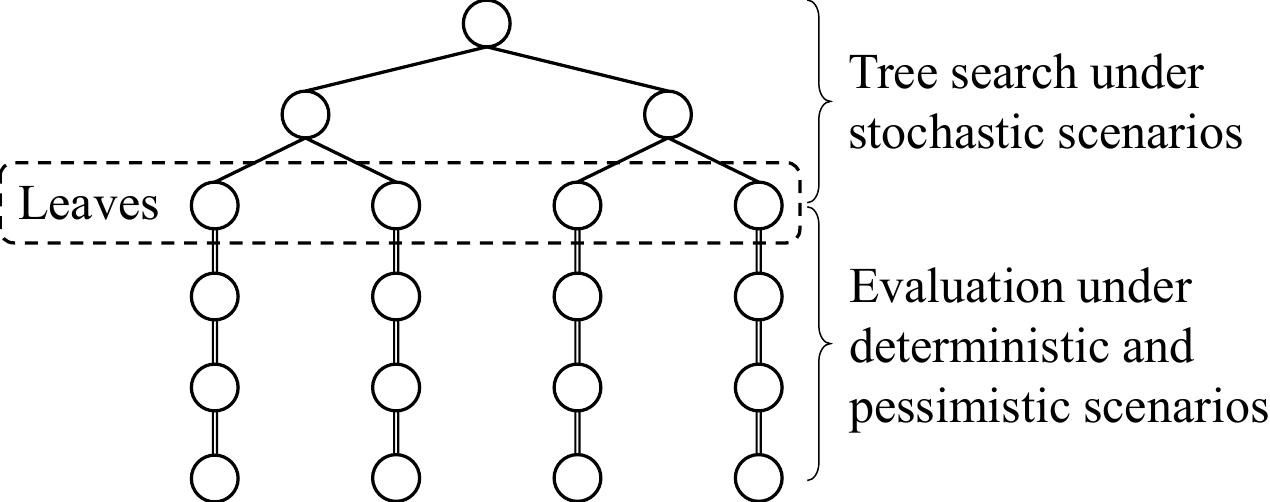}
  \caption{Tree search with deterministic and pessimistic scenarios.}
  \label{fig:tree}
\end{figure}

In our approach, the tree search after a specified depth is performed
under a deterministic and pessimistic scenario, which will also be
simply referred to as a deterministic scenario or a pessimistic
scenario, depending on what feature of the scenario is more relevant
in the context.  Figure~\ref{fig:tree} shows an example.  Here, the
tree search is performed in a standard manner until the depth of 2.
In this example, the branching factor is 2, and there are 4 nodes at
the depth of 2.  From each of these 4 nodes, ``tree search'' is
continued until it reaches the depth of 5 by assuming a deterministic
scenario.  Because the scenario is deterministic, there are no
branches after the depth of 2.

One may also interpret our approach as evaluating the 4 leaves (at the
depth of 2) on the basis of the deterministic scenario from the depth
of 2 to the depth of 5, and our following discussion will be based on
this interpretation.  Our approach keeps the size of the search tree
small, because there are branches only until a limited depth $L$.  At the
same time, our approach can take into account the events that can occur far ahead in the future,
because the leaves (nodes at depth $L$) can be evaluated with a
deterministic scenario that can be much longer than what would be
possible with branches.  What differs from the rollout in MCTS is that
we let the deterministic scenario be pessimistic, as we will discuss
in Section~\ref{sec:tree:pessimistic}.

More specifically, our approach performs standard tree search ({\it
  e.g.}, MCTS, exhaustive tree search, etc.) but on the search tree of
a limited depth $L$, evaluating the leaves (nodes at the depth $L$)
with pessimistic scenarios, and select the best action for the root
node ({\it e.g.}, as is selected with minimax tree search or with
multi-player tree search such as max$^n$ tree search \citep{maxn},
paranoid search algorithm \citep{paranoid}, and best-reply search
\citep{BestReplySearch}).  Namely, {\em our approach can use any tree
search algorithms for multi-player games but limit the depth of the
search tree at a given $L$ and evaluate the leaves with pessimistic
scenarios}.  One can choose the value of $L$ in consideration of
real-time constraints.

Algorithm~\ref{alg:pessimistic} shows the general framework of our
real-time tree search with pessimistic scenarios.  Notice that, except
at the depth of $L$, {\tt PessimisticTreeSearch} is nothing but the
standard tree search for games, and this tree search may also be done
non-exhaustively similar to MCTS.  For a node at depth $L$, {\tt
  PessimisticTreeSearch} evaluates the node under a pessimistic
scenario, which relies on two subroutines: {\tt PessimisticSimulation}
and {\tt Evaluate}.  Given a state of the game, {\tt
  PessimisticSimulation} generates a sequence of states from that
state in a pessimistic manner.  Although Pommerman is an imperfect
information game and an information set can be associated with a node,
we will associate a single state to the node by resolving
uncertainties, as we will discuss in
Section~\ref{sec:tree:pessimistic}.  Then the forward model of
Pommerman can be used to simulate the transitions of states.  Specific
procedures of {\tt PessimisticSimulation} will be described in
Section~\ref{sec:tree:pessimistic}.  {\tt Evaluate} gives the score of
the leaf on the basis of the sequence of the states given by {\tt
  PessimisticSimulation}.  Specific procedures of {\tt Evaluate} will
be described in Section~\ref{sec:tree:evaluate}.

\SetKwInput{KwInput}{Input}                
\SetKwInput{KwOutput}{Output}              

\begin{algorithm}[t]
  \DontPrintSemicolon
  \SetAlgoLined

  {\bf PessimisticTreeSearch}({\tt node})\;

  \KwInput{
    {\tt node} is the root node where tree search starts;
    $L$ is the depth of tree search under stochastic scenarios
  }
  \KwOutput{The best action at {\tt node}}

  \uIf{$\mbox{depth of } {\tt node} = L$}{
    \tcc{Evaluate {\tt node} under a pessimistic scenario}
    ${\tt board} \leftarrow {\tt GetState}({\tt node})$\;
    \tcp*{Get the game state at {\tt node}}

    ${\tt board\_sequence} \leftarrow {\tt PessimisticSimulation}({\tt board})$\;

    ${\tt score}({\tt node}) \leftarrow {\tt Evaluate}({\tt board\_sequence})$\;
  }\ElseIf{all children of {\tt node} have been evaluated}{
    \uIf{$\mbox{depth of {\tt node}} = 0$}{
      Find the bast action at {\tt node} based on the score of its children\;
    
      {\bf Return:} the best action at {\tt node}
    }\Else{
      \tcc{Compute the score of {\tt node} from the scores of its children}
      ${\tt score}({\tt node}) \leftarrow {\tt GetScore}({\tt node})$
    }
  }
  
  ${\tt node} \leftarrow {\tt FindUnevaluatedNode}()$\;
  \tcp*{e.g., via depth/breadth first search}
  
  {\tt PessimisticTreeSearch}({\tt node})
  \tcp*{Recursively run from {\tt node}}
  
  \caption{The general framework of the real-time tree search with pessimistic scenarios}
  \label{alg:treesearch}
\end{algorithm}

More specifically, in the Pommerman agents that implement the proposed
approach ({\it i.e.}, HakozakiJunctions ({\tt hakozaki}) and
dypm-final ({\tt dypm})), the depth of tree search is set as $L=1$.
The {\tt hakozaki} agent considers all of the leaves at depth 1 but
might need to choose an action before exhaustively searching all of
the leaves due to timeout.  On the other hand, the {\tt dypm} agent
considers six leaves at depth $L=1$ by taking into account only its own
actions, and the effect of the the actions by the other agents are
taken into account in evaluation with the deterministic scenario.  In
both of {\tt hakozaki} and {\tt dypm}, the leaves are evaluated with a
deterministic scenario with the length of at least 10 to take into
account the explosion of bombs, whose lifetime is 10.  Recall that the
{\tt eisenach} agent can perform the tree search only at the average
depth of $L=2$.  Hence, if we performed the standard tree search for 2
steps, there would be no computational budget left for the evaluation
with deterministic scenarios.

\subsection{Generating pessimistic scenarios}
\label{sec:tree:pessimistic}

From each of the leaves in the search tree, we generate a
deterministic scenario.  A key idea in our approach is to make this
deterministic scenario be pessimistic, which we will discuss in this
section with specific instantiation as Pommerman agents.

A pessimistic scenario can be generated in a systematic manner as
follows.  We assume that the state of the environment can be
represented by the positions of objects.  In Pommerman, these objects
are agents, bombs, flames, power-up items, and walls.  Some of those
objects change their positions randomly or by depending on the actions
of the agents, which forces the search tree to have branches.  If one can
tell the worst sequence of the positions of an object among all of the
possibilities, one can place and move that object accordingly in the
pessimistic scenario.  It is often the case, however, that the worst
positions are unknown.  Instead, {\em we generate a pessimistic
  scenario by allowing the objects to be located at multiple
  positions} even if that is unrealistic or illegal.  In Pommerman,
this typically corresponds to assuming that an opponent takes multiple
actions simultaneously in a nondeterministic manner, which means that
the opponent is copied into multiple positions in the next step.
Notice that such generated pessimistic scenarios can be more adversarial than
assuming the worst possible scenarios, because an object cannot
actually be at multiple positions.

Algorithm~\ref{alg:pessimistic} shows an example of how a pessimistic
scenario is generated from the leaf (node at depth $L$) of the search
tree.  This {\tt PessimisticSimulation} samples a sequence of states
from the leaf node, similar to rollout in MCTS, but the sequence of
states is then made pessimistic by allowing an object to be placed at
multiple positions.

\begin{algorithm}[t]
  \DontPrintSemicolon
  \SetAlgoLined
  
  \KwInput{
    {\tt leaf} is the leaf to evaluate;
    {\tt length} is the length of a pessimistic scenario;
    {\tt pessimism\_level} is the level of pessimism;
  }
  \KwOutput{The sequence of states from the leaf under a pessimistic scenario}

  ${\tt boards}[0] \leftarrow {\tt GetState}({\tt leaf})$\;
  \tcp*{Get the state of the game at {\tt leaf}}
    
  \tcc{Sample a sequence of states from the given state}
  \For{$\ell=1,\ldots,{\tt length}$}{
    ${\tt boards}[\ell] \leftarrow {\tt ForwardModel}({\tt boards}[\ell-1])$\;
    \tcp*{One of next states of game}
  }
  
  \tcc{Make the sequence of states pessimistic}
  \For{$\ell=1,\ldots,{\tt pessimism\_level}$}{    
    \For{each {\tt object} in ${\tt boards}[\ell-1]$}{
      {\tt positions} $\leftarrow$ Find the next positions of {\tt object} other than the one in ${\tt boards}[\ell]$\;

      \For{each {\tt position} in {\tt positions}}{
        Copy and place {\tt object} at {\tt position} in ${\tt boards}[\ell]$
      }
    }  
  }
  
  {\bf Return:} {\tt boards}

  \caption{An example of {\tt PessimisticSimulation}, which generates
    a pessimistic scenario for the case where the state of the game
    can be represented by the positions of objects.}
  \label{alg:pessimistic}
\end{algorithm}

Algorithm~\ref{alg:pessimistic} involves two hyper-parameters: {\tt
  length} and {\tt pessimism\_level}.  Here, {\tt length} denotes the
length of the pessimistic scenario used to evaluate the given leaf.
One can choose the value of {\tt length} in consideration of real-time
constraints.  On the other hand, {\tt pessimism\_level} denotes the
level of pessimism and controls how pessimistic the pessimistic
scenario is.  The value of {\tt pessimism\_level} can be optimized via
self-play in a way that the overall performance is maximized, which
will be further discussed in the following.

What is essential is that a pessimistic scenario is deterministic to
avoid the computational complexity resulting from branching in the
search tree.  Our guideline is to make this deterministic scenario
rather pessimistic, because good actions are often the ones that
perform well under pessimistic scenarios particularly in cases where safety is a
primary concern.  In Pommerman, an agent dies if it cannot escape from
flames, and our team loses if both of our agents die.  It is thus of
critical importance to ensure that our agents can survive,
while attacking opponents or collecting power-up
items.

In Pommerman, a deterministic scenario is represented by a sequence of
boards, where each board is given by the state of the game at a
certain point in time.  Such a deterministic scenario can be generated
by a forward model of Pommerman after resolving uncertainties.  There
are two sources of uncertainties: the future actions of agents and
partial observability.  These uncertainties can be resolved in
arbitrary ways, but our guideline is to resolve them in rather
pessimistic manners and to optimize the level of pessimism by tuning
hyperparameters via self-play.  A caveat is that the purpose of a
pessimistic scenario is not to find a proper lower bound on the value
of a leaf but to find a good action as a result of
evaluating that leaf with the pessimistic scenario.  Therefore, a
pessimistic scenario can be more adversarial or less adversarial than
the worst scenario.  The self-play allows us to optimize the level of
pessimism.

More specifically, both of {\tt hakozaki} and {\tt dypm} agents
generate a sequence of boards by letting the other agents move to
multiple positions simultaneously.  They then record the time when
each position is first occupied by an agent.  There are differences
between {\tt hakozaki} and {\tt dypm} in exactly what information is
recorded in the sequence of boards.  In {\tt hakozaki}, each position
in the $t$-th board has the information about when that position was
occupied, if the position has been occupied by the $t$-th step.  In
{\tt dypm}, each position in the $t$-th board has the information
about whether that position has been occupied by the $t$-th step.
Also, {\tt dypm} assumes that the other agents take actions only for a
predetermined number of steps (a hyperparameter tuned via self-play),
while {\tt hakozaki} does not have this limit.

Note that the sequence of such boards is in general illegal or
unrealistic.  There may be multiple copies of an agent in the board
({\tt dypm}), and an agent that might occupy a position may be
replaced by the integer value representing when that position can be
occupied ({\tt hakozaki}).  Also, some of the uncertainties are
resolved in a way that is not necessarily pessimistic.  For example,
we ignore the possibility that an agent might kick bombs in the
sequence of boards.  However, the purpose of the sequence of boards is
not to compute a proper lower bound of the value but to quickly
estimate the relative values of the leaves in the search tree in a way
that it gives the overall best performance.  Note that the part of
tree search (with branches) can take into account all of
the details, unlike the part of evaluation with a deterministic
scenario.  For example, if the action of an agent within the tree
search is to kick a bomb (by moving to the bomb), the movement of the
kicked bomb is taken into account in the remaining tree search as well
as in the deterministic scenario.  Our approach of tree search with a
pessimistic scenario allows us to take into account all of the details
in the near future (via tree search) as well as possibly critical
events in the distant future (via a pessimistic scenario).

\subsection{Evaluation with pessimistic scenarios}
\label{sec:tree:evaluate}

This sequence of boards is then used to estimate the value of the
initial board in the sequence ({\it i.e.}, a leaf).
If we obtain reward sufficiently frequently, we could
use Monte Carlo return (cumulative reward obtained during the sequence
of boards) as an estimate of the value.  However, in Pommerman, reward
is obtained only at the end of a game, and the Monte Carlo return has
very high variance.  We thus design the value in a way that choosing
actions that give high value tends to eventually achieving the goal.
The goal of a Pommerman agent is to knock down all of the opponents,
while the agent or its teammate is surviving.  The value should thus
reflect some notion of the survivability of the agent itself, its
teammate, and its opponents.  Roughly speaking, high value should
imply high survivability of the agent itself and its teammate, low
survivability of the opponents, or both.

In Pommerman, the survivability of an agent can be captured by the
number of positions that the agent can stay safely in the sequence of
boards.  More specifically, given a deterministic scenario, {\tt dypm}
counts the number of the time-position pairs from which an agent can
survive at least until the end of the scenario.  This number is used
as the survivability of the agent.  Namely, the survivability of the
agent is computed by first searching the reachable time-position pairs
in the sequence of boards and then pruning those pairs from which one
cannot survive until the end of the sequence.  On the other hand, {\tt
  hakozaki} finds the positions that an agent can reach at the end of
the sequence of boards, and compute the survivability on the basis of
the integer values that represent when the positions might be occupied
by other agents.  Intuitively, an agent is considered to have high
survivability if there are many positions that the agent can reach
without contacting the other agents.

Note that an agent $i$ computes the survivability $S(j,s)$ for each
leaf (state) $s$ and for each agent $j$ who is visible from $i$,
including $i$ itself.  The survivability of an agent $j\neq i$ is
computed on the basis of the sequence of boards that is pessimistic to
$j$ ({\it i.e.}, the agents except $j$ move to multiple positions
simultaneously)\footnote{In {\tt dypm}, to save computational cost, a
  single sequence of boards with no move of agents is used to compute
  the survivabilities of agents $j\neq i$, and those survivabilities
  are normalized by dividing them by the corresponding survivabilities
  when the agent $i$ does not exist.}.

Now, one can choose the best action on the basis of these
survivabilities.  Roughly speaking, our agent chooses the action that
maximizes the product the survivabilities of the agent
itself\footnote{To be more aggressive, the {\tt dypm} agent clips its
  own survivability $S$ at a threshold $S_{\rm th}$ when $S$ exceeds
  $S_{\rm th}$, where $S_{\rm th}$ is tuned via self-play.  } and its
teammate divided by the product of the survivabilities of the
opponents.  Because the survivability is defined for each leaf, which
corresponds to a combination of the actions of all agents, one needs
to marginalize out the actions of the other agents to define the
survivability of an agent with a particular action.  The survivability
of an agent when that agent takes a particular action can be defined
to be the minimum survivability of that agent given that the agent
takes that action ({\it i.e.}, worst case).  The survivability of a
teammate can also be defined as the minimum survivability.  On the
other hand, we find that the survivability of an opponent with that
action should be defined as the average survivability rather than
minimum or maximum.  These are how the survivabilities with a
particular action are defined in {\tt hakozaki}.  On the other hand,
each leaf in the search tree of a {\tt dypm} agent corresponds to each
action of that agent, and there is no need for marginalization.  A
caveat is that the action by the {\tt dypm} agent might be blocked by
other agent, resulting in no move.  The survivability with such an
action is thus averaged with the survivability of no move.

\makeatletter

In this section, we have discussed how our agents choose actions in
most critical situations of Pommerman, where the agents interact with
other agents.  In those situations, the goal of an agent is to reduce
the survivabilities of the opponents, while keeping the
survivabilities of the agent itself or its teammate sufficiently high.
In other situations, the agent can ignore the behavior of other agents
and seek to attain intermediate objectives.  Examples of such
objectives include (\@roman{1}) breaking wooden walls to make passages
or to find power-up items, (\@roman{2}) collecting power-up items, and
(\@roman{3}) moving towards the areas that cannot been observed to
obtain new information.  Although there is a question of what
objective to pursue at each step, it is relatively easy to choose
actions once an objective is given, because we do not need to take
into account the actions of the other agents.  Our agent heuristically
sets an objective and chooses actions by the use of a standard search
technique until the agent meets and starts interacting with other
agents.

\makeatother

\section{Experiments}

We conduct two sets of experiments to investigate the overall
performance of the proposed approach and the effectiveness of our key
idea ({\it i.e.}, the use of pessimistic scenarios in tree search).
Although the agents that implement our proposed approach have won the
first and third places in the NeurIPS 2018 Pommerman competition, the
number of matches in the competition is too limited to draw
conclusions.  In the first set of experiments, we intensively evaluate
the performance among the top five teams, from the competition, that
implement state-of-the-art approaches, including ours.  In the second
set of experiments, we study the effectiveness of pessimism by
changing one of the key hyperparameters of {\tt dypm} that controls
the level of pessimism in the proposed approach.

\subsection{Performance against state-of-the-art agents}

The competition was run according to a double
elimination style with two brackets, where the team that won two games
before the other moves on to the next round.  A tie was replayed for
the first time, but another tie was resolved by matches on another
version of Pommerman, where walls can collapse.  Namely, tie was not
the same as ``no game'' in the competition, and the results in this
section needs to be interpreted accordingly.  See
\citet{PommermanNeurIPS2018} for more details about the settings of
the competition.

The top five teams were {\tt hakozaki}, {\tt eisenach}, {\tt dypm},
{\tt navocado}, and
{\tt skynet}.
The top three teams are based on tree search, as we have discussed in Section~\ref{sec:tree}.
The other two are based on reinforcement learning.  More specifically, {\tt navocado}
is based on advantage-actor-critic \citep{Navocado},
and {\tt skynet} is based on proximal policy optimization.

Table~\ref{tbl:neurips} shows the results of the direct matches among
the top five teams in the competition.  For example, {\tt eisenach}
defeated {\tt dypm} four times, {\tt dypm} defeated {\tt eisenach}
once, and there were no ties between these two teams.  Although the
winners of the competition were determined according to the rule of the
competition, the number of matches was clearly
limited to statistically determine which teams are stronger than
others.  In particular, there were pairs of teams that had no direct
matches in the competition.

\begin{table}[tbh]
\centering
{\tt \footnotesize
\begin{tabular}{@{}rccccc@{}}
\toprule
& \hspace{-0.5mm}hakozaki$^\star$ & \hspace{-0.5mm}eisenach & \hspace{-0.5mm}dypm$^\star$ & \hspace{-0.5mm}navocado & \hspace{-0.5mm}skynet \\
\midrule
hakozaki$^\star$ & -     & 4/2/1 & -     & 2/0/2 & 2/0/0 \\
eisenach & 2/4/1 & -     & 4/1/0 & -     & -     \\
dypm$^\star$     & -     & 1/4/0 & -     & 2/1/0 & -     \\
navocado & 0/2/2 & -     & 1/2/0 & -     & 2/1/5 \\
skynet   & 0/2/0 & -     & -     & 1/2/5 & -     \\
\bottomrule\\
\end{tabular}
}
\caption{A summary of the matches among the top five teams in the
  competition.  Each entry shows the number of
  ``wins / losses / ties'' for a row agent against a column agent.
  The $\star$ marks indicate the teams that implement our
  approach.}
\label{tbl:neurips}
\end{table}

The purpose of the following experiments is to complement the 
competition by running a greater number of matches
between the top five teams.  We use the Docker images\footnote{The
  Docker images are available as
  multiagentlearning/\{hakozakijunctions, eisenach, dypm.1, dypm.2,
  navocado, skynet955\}.  Note that the {\tt dypm} team uses two
  agents, dypm.1 and dypm.2, that differ only in the values of their
  hyperparameters.  In the other teams, the two agents are identical.}
of the agents that have been used in the
competition.  We run our experiments on a Ubuntu 18.04 machine having
eight Intel Core i7-6700K CPUs running at 4.00~GHz and 64~GB random
access memory.  Note that these computational resources are 
different from what has been used at the competition (exact
computational resources used at the competition are not known).
Therefore, the results of our experiments should not be considered as
a refinement but rather as complementary to the results from the
competition.

Table~\ref{tbl:exp} summarizes the results of the 1,000 matches that we
have run between each pair of the teams.  Because there are two
essentially different configurations for the initial positions of the
two teams, half of the matches are initialized with one configuration,
and the other with the other configuration.  Each team is also matched
against itself for 500 matches, and the results for both sides are
counted in the table (if a match is tied, two ties are counted).  In
total, we run 22,500 matches, which take approximately two weeks in
our environment.

\begin{table*}[tbh]
  \centering
  {\tt \footnotesize
\begin{tabular}{@{}rrrrrrrrrrrrrrrr@{}}
\toprule
& hakozaki$^\star$ & eisenach & dypm$^\star$ & navocado & skynet & TOTAL \\
\midrule
hakozaki$^\star$ & 112/112/776 & 490/139/371 & 279/221/500 & 694/  3/303 & 845/  7/148 & 2420/482/2098\\
 eisenach       & 139/490/371 & 338/338/324 & 403/492/105 & 866/ 85/ 49 & 918/ 55/ 27 & 2664/1460/876\\
    dypm$^\star$ & 221/279/500 & 492/403/105 & 107/107/786 & 935/ 12/ 53 & 969/  0/ 31 & 2724/801/1475\\
 navocado       &   3/694/303 &  85/866/ 49 &  12/935/ 53 &  95/ 95/810 & 198/ 50/752 &  393/2640/1967\\
skynet955       &   7/845/148 &  55/918/ 27 &  0/969/ 31  &  50/198/752 &  57/ 57/886 &  169/2987/1844\\
\bottomrule\\
\end{tabular}
}
  \caption{ A summary of the 1,000 matches between each pair among the
    top five teams.  Each entry shows the number of ``wins / losses /
    ties'' for a row agent against a column agent.  The rightmost
    column shows the total number of ``wins / losses / ties'' for row
    agents.  The $\star$ mark indicates the team that implements the
    proposed approach.}
  \label{tbl:exp}
\end{table*}

In our experiments, the top three teams ({\tt hakozaki}, {\tt
  eisenach}, {\tt dypm}) have completely dominated the other two ({\tt
  navocado}, {\tt skynet}), recording 5,227 wins (87.1~\%), 162 losses
(2.7~\%), and 611 ties (10.2~\%).  In particular, {\tt hakozaki} and
{\tt dypm}, who implement our proposed approach, have lost
against {\tt navocado} or {\tt skynet} only in 22 matches (0.4~\%).  While the top three teams are
based on tree search, the other two are based on model-free
reinforcement learning.  This indicates the advantages of tree search
in Pommerman, where precise planning in the following several steps is
critically important to survive from the explosion of bombs.  The
results in our experiments are not necessarily consistent with those
in the competition (Table~\ref{tbl:neurips}), however.  In particular,
{\tt dypm} has lost once against {\tt navocado} in the competition.
This may be because the {\tt dypm} agents have occasionally
experienced timeouts ({\it i.e.}, the agent does not respond in 100
milliseconds, which is treated as a ``stop'' action) in the
competition, because {\tt dypm} does not implement any mechanisms for
measuring the elapsed time.  Also, the computational resources in our
experiments might be less favorable to the learning agents ({\tt
  navocado} and {\tt skynet}) than what has been used in the
competition.

Among the top three teams, {\tt hakozaki} has consistently dominated
the others, although the relative advantages between {\tt hakozaki} and {\tt dypm}
are relatively small.  Also, {\tt dypm} has slightly outperformed {\tt
  eisenach}.  Overall, {\tt hakozaki} and {\tt dypm}, who implement
the proposed approach, have recorded 982 wins (49.1~\%), 542 losses
(27.1~\%), and 476 ties (23.8~\%) against {\tt eisenach}.  Note that
these top three teams implement their agents and forward models in
different programming languages: {\tt hakozaki} in Java\texttrademark, {\tt
  eisenach} in C++, and {\tt dypm} in Python.  Also, the {\tt dypm}
agent runs on a single thread, while {\tt hakozaki} and {\tt eisenach}
use multiple threads.  Overall, our experimental results suggest the
superiority of the proposed approach in real-time tree search for
sequential decision making with limited computational resources.

\subsection{Effectiveness of pessimism}

We next study the effectiveness of the pessimism in the deterministic
scenario used in our proposed approach.  Recall that a {\tt dypm}
agent generates a deterministic scenario by assuming that the other
agents take multiple actions simultaneously in a nondeterministic
manner but only for a limited number of steps.  Here, we refer to this
limited number of steps as the pessimism level and study how the
performance of {\tt dypm} depends on the pessimism level.
Specifically, for each pessimism level, we run 1,000 matches
against a baseline and record the number of wins, losses, and times.
The baseline is either a team of default agents ({\tt
  SimpleAgent}) or a team of the {\tt dypm} agents whose 
pessimism level is set 0.

\begin{figure}[t]
  \begin{minipage}{0.49\linewidth}
    \centering
    \includegraphics[width=\linewidth]{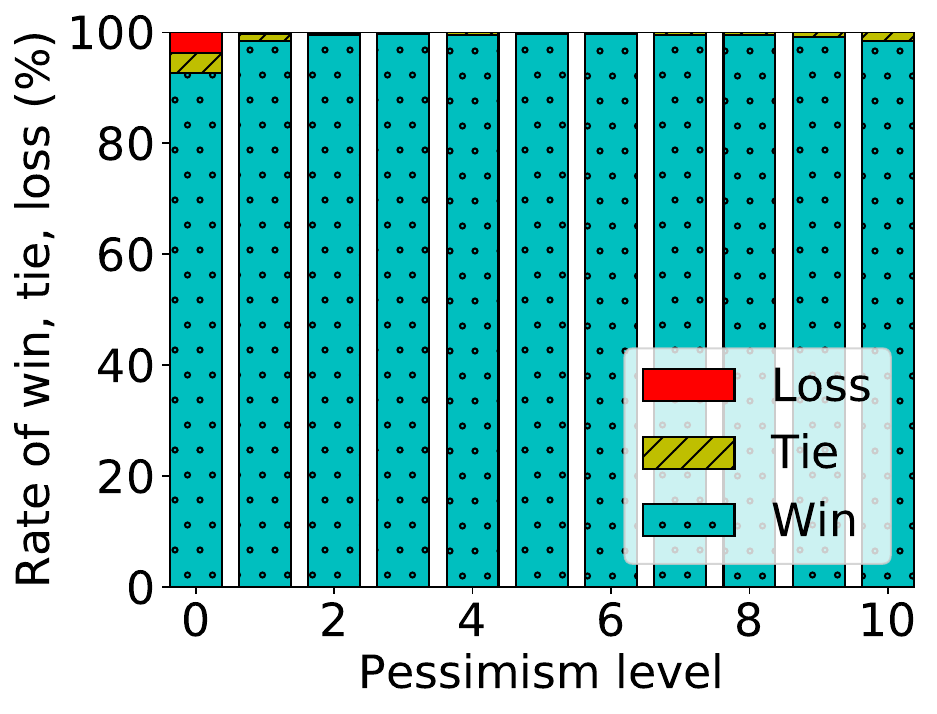}\\
    (a) vs. {\tt SimpleAgent}
  \end{minipage}
  \begin{minipage}{0.49\linewidth}
    \centering
    \includegraphics[width=\linewidth]{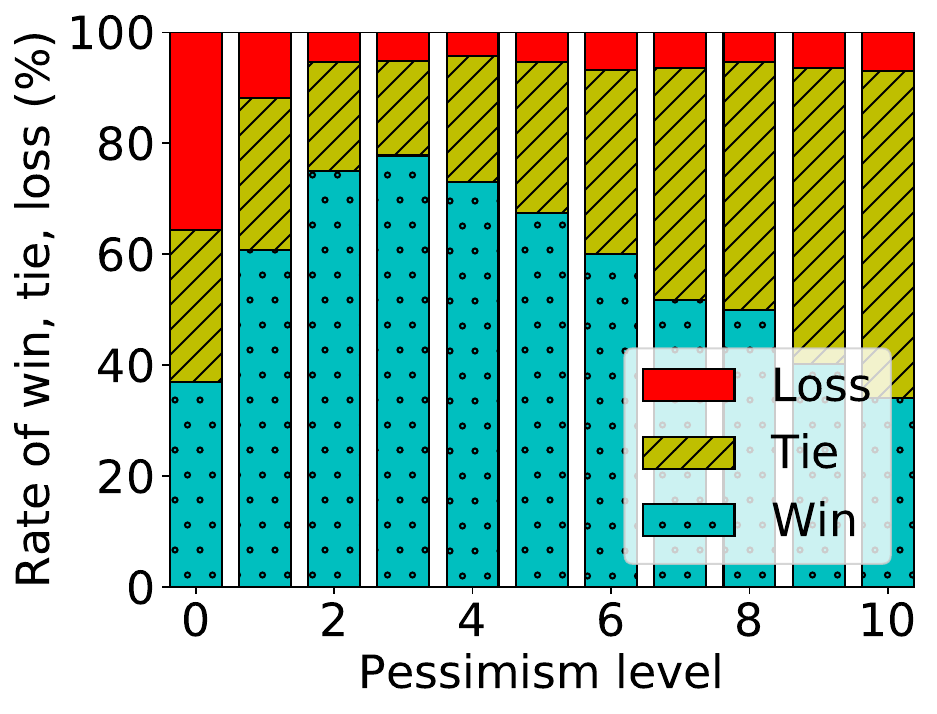}\\
    (b) vs. no pessimism
  \end{minipage}
  \caption{The performance of {\tt dypm} with varying pessimism
    levels.  The rate of wins, losses, and ties of {\tt dypm} against
    a baseline is shown for each pessimism level (from 0 to 10) as
    indicated along the horizontal axis.  The baseline is {\tt
      SimpleAgent} in (a) and the {\tt dypm} with no pessimism (level
    0) in (b).}
  \label{fig:pessimism}
\end{figure}

Figure~\ref{fig:pessimism} shows the number of wins, losses, and ties
of {\tt dypm} against each baseline, where the pessimism level in {\tt
  dypm} is varied as indicated along the horizontal axis.  For example,
{\tt dypm} with the pessimism level 3 has had 997 wins, 1 loss, and 2
ties against {\tt SimpleAgent} (Figure~\ref{fig:pessimism}~(a)), and
this pessimism level is the one that has actually been set in {\tt
  dypm} in the competition.  The winning rate of {\tt dypm} against
{\tt SimpleAgent} increases from 92.6~\% to 99.7~\% by increasing the
pessimism level from 0 to 3.  Further increasing the pessimism level
can reduce the number of losses but increases the number of ties, but
these changes are insignificant with the very high winning rate.

To clarify these changes, Figure~\ref{fig:pessimism}~(b) shows the
results against a stronger baseline, which is {\tt dypm} with
pessimism level 0.  Now, the winning rate of {\tt dypm} against the
baseline increases from 36.9~\% to 77.8~\% by increasing the pessimism
level from 0 to 3.  Further increasing the pessimism level from 3 to 4
can reduce the rate of losses from 5.1~\% to 4.2~\% but increases the
rate of ties from 17.1~\% to 22.7~\%.

Overall, we find that pessimism in deterministic scenarios can
significantly improve the overall performance of Pommerman agents.  Also,
the performance is sensitive to the particular level of pessimism, and
an appropriate level of pessimism may be determined via self-play.
Note, however, that Pommerman is an extensive-form game with imperfect
information, and the optimal strategy in general is to
probabilistically mix multiple levels of pessimism.  Also, {\tt dypm}
has other hyperparameters, and the optimal level of pessimism depends
on the values of the other hyperparameters.

\section{Conclusion}

We have proposed an approach of real-time tree search, where tree
search is performed only with a limited depth, and the leaves are
evaluated under a deterministic and pessimistic scenario.  Because
there is no branching with a deterministic scenario, our evaluation
can take into account the events that can occur far ahead in the
future.  Also, evaluation with pessimistic scenarios can lead to
selecting good actions, which are often the ones that perform well
under the pessimistic scenarios particularly in cases where safety is
a primary concern.  We have assumed that the state can be represented
by the positions of objects and generated pessimistic scenarios by
allowing the objects to be located at multiple positions even if that
may be unrealistic or illegal.  One could, however, apply the general
idea of our real-time tree search with pessimistic scenarios to a
broader range of domains by designing pessimistic scenarios suitable
for particular domains.  For example, for applications to autonomous
robots, drones, or other agents in continuous space, a
pessimistic scenario may be generated by assuming objects (e.g., other
robots and drones) that increases the size over time.

Our experiments with Pommerman suggest that the performance of the
proposed approach is sensitive to the particular level of pessimism,
but it can be optimized via self-play.  With the optimized level of
pessimism, the proposed approach is shown to outperform other
state-of-the-art approaches to real-time sequential decision making.
An interesting direction of future work is to combine the proposed
approach with model-free reinforcement learning, where the proposed
approach is used to choose specific actions in each step to attain the
intermediate objective that is selected by model-free reinforcement learning.
Such integration of tree search and reinforcement learning
is an active field of research \citep{AlphaZero,ATB17,VSS17,JEL18,EDSM19,PGSearch,SXX19}.

\section*{Acknowledgments}
Takayuki Osogami was supported by JST CREST Grant Number JPMJCR1304, Japan.

\bibliographystyle{named}
\bibliography{pommerman,RL}

\end{document}